\documentclass{article}

\PassOptionsToPackage{numbers, compress}{natbib}

\usepackage[preprint]{neurips_2022}

\usepackage[utf8]{inputenc} 
\usepackage[T1]{fontenc}    
\usepackage{hyperref}       
\usepackage{url}            
\usepackage{booktabs}       
\usepackage{amsfonts}       
\usepackage{nicefrac}       
\usepackage{microtype}      
\usepackage{xcolor}         

\usepackage{amsmath,amssymb,amsfonts}
\usepackage{algorithmic}
\usepackage{algorithm}
\usepackage{graphicx}
\usepackage{textcomp}
\usepackage{xcolor}
\usepackage{caption}
\usepackage{subcaption}

\newtheorem{theorem}{Theorem}

\usepackage{bbm}

\title{FEAMOE: Fair, Explainable and Adaptive Mixture of Experts}

\author{%
  Shubham Sharma \\
  University of Texas at Austin\\
  \texttt{shubham\_sharma@utexas.edu} \\
  \And
  Jette Henderson \\
  CognitiveScale\\
  \texttt{jhenderson@cognitivescale.com} \\
  \And
  Joydeep Ghosh \\
  University of Texas at Austin\\
  \texttt{jghosh@utexas.edu} \\
}

\begin{document}

\maketitle

\begin{abstract}
  Three key properties that are desired of trustworthy machine learning models deployed in high-stakes environments are fairness, explainability, and an ability to account for various kinds of "drift". While drifts in model accuracy, for example due to covariate shift, have been widely investigated, drifts in fairness metrics over time remain largely unexplored. In this paper, we propose FEAMOE, a novel "mixture-of-experts" inspired framework aimed at learning fairer, more explainable/interpretable models that can also rapidly adjust to drifts in both the accuracy and the fairness of a classifier. We illustrate our framework for three popular fairness measures and demonstrate how drift can be handled with respect to these fairness constraints. Experiments on multiple datasets show that our framework as applied to a mixture of linear experts is able to perform comparably to neural networks in terms of accuracy while producing fairer models. We then use the large-scale HMDA dataset and show that while various models trained on HMDA demonstrate drift with respect to both accuracy and fairness, FEAMOE can ably handle these drifts with respect to all the considered fairness measures and maintain model accuracy as well. We also prove that the proposed framework allows for producing fast Shapley value explanations, which makes computationally efficient feature attribution based explanations of model decisions readily available via FEAMOE.
\end{abstract}

\section{Introduction}

The field of responsible artificial intelligence has several desiderata that are motivated by regulations such as the General Data Protected Regulation \citep{butterworth2018ico}. These include: ensuring that an AI model is non-discriminatory and transparent; individuals subject to model decisions should have access to explanations that point a path towards recourse; and models should adapt to any changes in the characteristics of the data post-deployment so as to maintain their quality and trustworthiness.

Most approaches towards the mitigation of any form of bias assume a static classifier. A practitioner decides on some definition of fairness, trains a model that attempts to enforce this notion of fairness and then deploys the model. Many of the fairness definitions are based on model outcomes or on error rates (the gap between true and/or false positive rates) that are associated with different subgroups specified by a protected attribute. The goal is to reduce the difference between these error rates across relevant subgroups. For example, average odds difference \citep{bellamy2018ai} is a measure signifying equalized odds and is given by the sum of the differences in both true positive and false positive rates between two groups, scaled by a factor of 0.5. Equality of opportunity and demographic parity \citep{barocas-hardt-narayanan} are also popular definitions of fairness. Recently, fairness in terms of a gap of recourse has been proposed, where recourse is defined as the ability to obtain a positive outcome from the model \citep{sharma2020fair}. While the suitability of a fairness measure is application dependent \citep{mehrabi2019survey,barocas-hardt-narayanan}, demographic parity and equalized odds remain the most popularly used, and the need for recourse gap-based fairness is being increasingly recognized \citep{karimi2020survey}.

However, static models can encounter drift once deployed, as the statistical properties of real data often change over time. This can lead to deteriorating performance. Model drift can occur when the properties of the target variable change (concept drift) or when the input data distribution changes, or both. The performance of models has largely been measured through accuracy-based metrics such as misclassification rates, F-score or AUC. \citep{stanley2003learning}.
However, a model trained in the past and found to be fair at training time may act unfairly for data in the present. Addressing drift with respect to fairness in addition to accuracy has remained largely unexplored though it is an important aspect of trustworthy AI in practice.

Explainability of individual model outcomes is another principal concern for trustworthy ML.  Among many methods of explanations in terms of feature attribution, \citep{bhatt2020explainable}, the SHAP approach based on Shapley values is particularly popular as it enjoys several axiomatic guarantees \citep{lundberg2017unified}. While computation of SHAP values is fast for linear and tree-based models, it can be very slow for neural networks and several other model types, especially when the data has a large numbers of features or when a large number of explanations are required \citep{molnar2019}. This poses a barrier to deployments that demand  fast explanations in real-time, production settings.

In this paper, we address these fairness, data/model drift, and explainability concerns by proposing FEAMOE: Fair, Explainable and Adaptive Mixture of Experts, an incrementally grown mixture of experts (MOE) with fairness constraints. In the standard mixture of experts setup, each expert is a machine learning model, and so is the gating network. The gating network learns to assign an input-dependent weight $g_u(\textbf{x})$ to the $u^{th}$ expert for input $\textbf{x}$, and the final output of the model is a weighted combination of the outputs of each expert. Hence, each expert contributes differently for every data point towards the final outcome, which is a key difference from standard ensembles.


Many types of MOE's exist in the literature \citep{yuksel2012twenty} - the architecture is not standard.
For FEAMOE, we chose this family, with some novel modifications described later, for three main reasons: 1) Suitable regularization penalties that promote fairness can be readily incorporated into the loss function. 2) Online learning is possible, so changes in the data can be tracked. Crucially, since localized changes in data distribution post-deployment may impact only one or a few experts, the other experts may not need to be adjusted, making the experts localized and only loosely coupled. This allows for handling drift and avoiding catastrophic forgetting, which is a prime concern in widely used neural network models \citep{robins1995catastrophic}. 3) Simpler models can be used to fit a more complex problem in the mixture of experts, as each model needs to fit well in only a limited part of the input space. In particular, even linear models, which provide very fast SHAP explanations, can be used. The overall mixture of experts, even with such simple base models (the "experts") often has predictive power that is comparable to a single complex model such as a neural network, as shown by our experiments as well as in many previous studies \citep{yuksel2012twenty}.

\begin{figure*}
     \centering
     \begin{subfigure}[b]{0.24\textwidth}
         \centering
         \includegraphics[width=\textwidth]{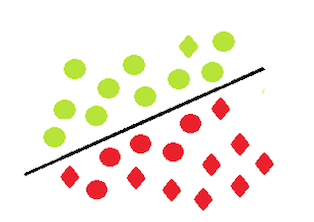}
         \caption{}
         \label{fig:1a}
     \end{subfigure}
     \hfill
     \begin{subfigure}[b]{0.24\textwidth}
         \centering
         \includegraphics[width=\textwidth]{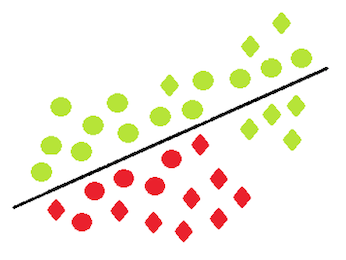}
         \caption{}
         \label{fig:1b}
     \end{subfigure}
     \hfill
     \begin{subfigure}[b]{0.24\textwidth}
         \centering
         \includegraphics[width=\textwidth]{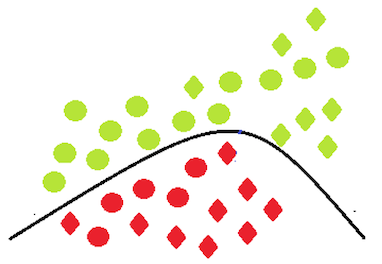}
         \caption{}
         \label{fig:1c}
     \end{subfigure}
          \hfill
     \begin{subfigure}[b]{0.24\textwidth}
         \centering
         \includegraphics[width=\textwidth]{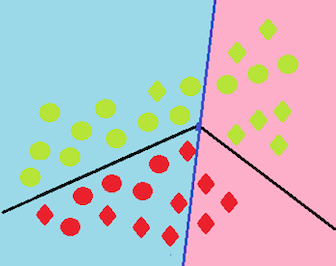}
         \caption{}
         \label{fig:1d}
     \end{subfigure}
        \caption{A toy example demonstrating the need and use of FEAMOE. The color of every datapoint corresponds to the original class label. Diamonds represent the underprivileged group and circles represent the privileged group. (a) Represents a perfectly accurate linear classifier, (b) represents the same classifier mis-classifying new data points and inducing bias (drift), (c) represents an alternate non-linear model that corrects for drift but has a complex decision boundary and (d) represents FEAMOE where the blue and pink regions show the regions of operation for each of the two experts, separated by the gating network}
        \label{fig:tOYmOE}
\end{figure*}

A motivating toy example of why FEAMOE is needed and how it works is shown in Figure \ref{fig:tOYmOE}. Consider a linear binary classifier (\ref{fig:1a}) that has perfect accuracy. The colors represent the ground truth labels, and green is the positive (desired) class label. The circles are the privileged group and diamonds are the underprivileged group. As can be seen in the figure, more diamonds receive a negative outcome and more circles receive a positive outcome. Consider new data that arrives for predictions. This classifier (\ref{fig:1b}) not only misclassifies individuals but also gives more underprivileged individuals that were actually in the positive class a negative outcome, hence inducing bias with respect to equalized odds. There is drift with respect to accuracy and fairness.
A more complex model (\ref{fig:1c}) such as a neural network, if retrained, may handle some of these concerns but would be less explainable.

FEAMOE can address these imperative concerns, as shown in \ref{fig:1d}. 
Trained in an online manner, a new linear model is added (i.e., an expert) once the new data arrives. 
The gating network dictates which region each expert operates in (shown by the blue and pink colors), and FEAMOE is able to adapt automatically with respect to accuracy and fairness. 
This dynamic framework enables the overall model to be fairer, adjust to drift, maintain accuracy, while also remaining explainable since the decision boundary is locally linear.

We show how three fairness constraints--demographic parity, equalized odds, and burden-based fairness--can be incorporated into the mixture of experts training procedure in order to encourage fitting fairer models (according to these measures). We use these three popular fairness measures as illustrative examples to demonstrate the effectiveness of FEAMOE, but our method can be adapted to incorporate other fairness constraints as well. 
We then describe a new algorithm for training to account for drift, where the drift in question can be due to accuracy or fairness.
We show experimentally that by using a set of logistic regression experts, the accuracy of the mixture is comparable to using a complex model like a neural network. Additionally, we show we can efficiently compute Shapley value explanations when explanations for every individual expert can be computed quickly. To the best of our knowledge, this is the first work that addresses the problem of drift with respect to fairness in a large-scale real world dataset. We then introduce a framework that can flexibly adapt to drifts in both fairness and accuracy with the added benefit of delivering explanations quickly, while comparing to the less explainable neural network model class trained in online mode.

The key contributions of this work are: a mixture of experts framework that can incorporate multiple fairness constraints, a method to handle drift, where drift can be with respect to accuracy or fairness, empirical evidence of the presence of drift with respect to fairness in a real-world, large-scale dataset, a theoretical proof that FEAMOE leads to the generation of fast explanations given a suitable choice of experts, and extensive experimentation on three datasets to show that our method has predictive performance similar to neural networks while being fairer, handling different types of drift, and generating faster explanations.

\section{Related Work}

The mixture of experts (MOE) \citep{jacobs1991adaptive, xu1994alternative} represent a class of co-operative ensemble models; detailed surveys on their design and use can be found in \citep{yuksel2012twenty} and \citep{masoudnia2014mixture}. 
Very recently, the deep learning community has started recognizing and leveraging several advantageous properties that MOE's have for efficient design of complex, multi-purpose learners \citep{riquelme2021scaling}. This paper contributes to this expanding literature by proposing a new algorithm to train this model class to account for both fairness and drift, and by also adding an explainability module.

Fairness in machine learning is a growing field of research \citep{hacker2018teaching}. Mitigating biases in models can be done through pre-processing, in-processing, or post-processing techniques. A description of these techniques can be found in \citep{bellamy2018ai}.  In-processing techniques for fairness have been gaining traction \citep{zhang2018mitigating,mehrabi2019survey, sharma2020fair}. However, there is limited work on investigating the usefulness of ensemble models in dealing with biases. \citep{grgic2017fairness} show that an ensemble of fair classifiers is guaranteed to be fair for several different measures of fairness, an ensemble of unfair classifiers can still achieve fair outcomes, and an ensemble of classifiers can achieve better accuracy-fairness trade-offs than a single classifier. However, they neither provide experimental evidence nor discuss specific methods to incorporate fairness into ensemble learning. \citep{madras2017predict} develop a method to learn to defer in the case of unfair predictions. \citep{bhaskaruni2019improving} use an AdaBoost framework to build a fairer model. \citep{nejdl2021farf} use adaptive random forest classifiers to account for fairness in online learning, only considering the statistical parity definition of fairness.

Accounting for drift is a widely explored problem, and is now appearing in commercial products as well
(e.g. model monitoring is a key part of MLOPs) as ML solutions get deployed in business environments. Details on many such approaches can be found in \citep{gama2014survey,lu2018learning}. Among these approaches, the one that comes closest to ours is \citep{stanley2003learning} which uses a committee of decision trees to account for drift. However, ensuring fairness in the presence of drift remains an open problem. \citep{biswas2020ensuring} is a very recent work on achieving a fairer model by building a set of classifiers in the presence of prior distribution shifts. The method is built for a shift between the training and test distributions, and not for online learning.

There are many ways to explain a machine learning model \citep{burkart2021survey,molnar2019}.  In this paper, we focus on Shapley values-based explanations, which are widely used in practical applications \citep{bhatt2020explainable}. \citep{lundberg2018consistent} propose the computation of Shapley values for tree ensembles, which is a faster way to get Shapley values than through the more broadly applicable method, KernelShap \citep{lundberg2017unified}. We show that in FEAMOE, the Shap values for the overall model are just a data-dependent linear combination of the values from individual experts. Thus the mixture approach does not add any significant complexity to the computation of feature attribution scores.

\section{Theory}

We first summarize the original mixture of experts framework  and then describe the addition of fairness constraints. Then, we introduce the algorithm to detect and mitigate data drift when the data input is sequential (online learning). Thereafter, we show how using the proposed mixture of experts architecture leads to computing faster Shapley value explanations for the overall non-linear model. 

Mixture of Experts (MoE) \citep{jacobs1991adaptive} is a technique where multiple experts (learners) can be used to softly divide the problem space into regions. A gating network decides which expert to weigh most heavily for each input region. Learning thus consists of the following: 1) learning the parameters of individual learners and 2) learning the parameters of the gating network. Both the gating network and every expert have access to the input $\textbf{x}$. The gating network has one output $g_{i}$ for every expert $i$. The output vector is the weighted (by the gating network outputs) mean of the expert outputs:
  $  y(\textbf{x}) = \sum_{i=1}^{m}g_{i}(\textbf{x})y_{i}(\textbf{x})$.
Consistent with \citep{jacobs1991adaptive}, the error associated with training the mixture of experts for case $j$ for an accurate prediction is given by:
$    E_{acc}^j = -log \sum_{i} g_i^j e^{\frac{-1}{2}||\textbf{d}^j-\textbf{y}_i^j||^2}$
,where $\textbf{y}_i^j$ is the output vector of expert $i$ on case $j$, $g_i^j$ is the proportional contribution of expert $i$ to the combined output vector, and $\textbf{d}^j$ is the desired output vector.

\subsection{Fairness Constraints}

In this paper, we incorporate three diverse fairness definitions into the mixture of experts framework: demographic parity only depends on the model outcome, equalized odds is conditioned on the ground-truth label, and burden-based fairness depends on the distance of the input to the boundary. These three popular definitions have been chosen as illustrative metrics; our approach can be readily extended to several other fairness metrics as well. 

For simplicity, we consider a binary classification setting with a binary protected attribute (our approach readily extends to multi-class and multi-protected attribute problems, where a protected attribute is a feature such as race or gender). Let $y^j_i = 1$ be the positive outcome. Let $A=0$ and $A=1$ represent the underprivileged and privileged protected attribute groups, respectively. For a given dataset $D$, let $D_{ad}$ represent all individuals that belong to the protected attribute group $a$ and original class label $d$.

Statistical parity difference (SPD), which is a measure of demographic parity, measures the difference between the probability of getting a positive outcome between protected attribute groups \citep{bellamy2018ai,sharma2020data}. Let $D_0$ be the set of individuals in the underprivileged group and $D_1$ be the set of individuals in the privileged group. Inspired by \citep{slack2020fairness}, the associated penalty for demographic parity for case $j$ is:
$     E_{SPD}^j = \mathbbm{1}{[j \in D_0]}(1-\sum_{i} g_i \textbf{y}_i^j) +  \mathbbm{1}{[j \in D_1]}(\sum_{i} g_i \textbf{y}_i^j).$
The idea behind this term is that individuals belonging to the underprivileged group predicted as getting a negative outcome are assigned a higher penalty. Similarly, individuals belonging to the privileged group predicted to have a positive outcome are assigned a higher penalty, thereby encouraging an SPD value closer to zero.

Average odds difference (AOD), which is a measure of equalized odds, measures the difference in true and false rates between protected attribute groups. Details on the measure can be found in  \citep{bellamy2018ai,sharma2020data}. The associated penalty for equalized odds is:\\
$   E_{AOD}^j = \mathbbm{1}{[j \in D_{01}]}(1-\sum_{i} g_i \textbf{y}_i^j) + \mathbbm{1}{[j \in D_{11}]}(1-\sum_{i} g_i \textbf{y}_i^j) + \mathbbm{1}{[j \in D_{10}]}(\sum_{i} g_i \textbf{y}_i^j) + \mathbbm{1}{[j \in D_{00}]}(\sum_{i} g_i \textbf{y}_i^j). $
This term encourages the true and false positive rate gaps between groups to reduce by conditioning the indicator function on the ground truth label in addition to the protected attribute (as was in the demographic parity formulation).

Burden for a protected attribute group is a measure of the ability to obtain recourse for individuals in that group. As shown in \citep{sharma2020fair}, burden-based fairness can be calculated as:
$Burden = \quad \mathop{\mathbb{| \; E}}_{\textbf{x}|{\emph{A}=0}}[d(\textbf{x},\mathcal{B})] \; -  \mathop{\mathbb{E}}_{\textbf{x}|{\emph{A}=1}}[d(\textbf{x},\mathcal{B})] \;|$
,where $d(\textbf{x},\mathcal{B})$ represents the distance to the boundary for a given $\textbf{x}$ that is classified as being in the negative class. Then, the associated penalty for burden based fairness is:
$ E_{Burden}^j = \quad \mathop{\mathbb{| \; E}}_{\textbf{x}|{\emph{A}=0}}[d(\textbf{x},\mathcal{B})] \; -  \mathop{\mathbb{E}}_{\textbf{x}|{\emph{A}=1}}[d(\textbf{x},\mathcal{B})]| \; .$
 
 \begin{algorithm}[t]

 \caption{Learning FEAMOE}
  \begin{algorithmic}
   \label{alg:algorithmmoe}
  \STATE Inputs: data $X$, labels $Y$
  \STATE Hyperparameters: $k$, $\Delta \lambda_{1}$, $\Delta \lambda_{2}$, $\Delta \lambda_{3}$
  \STATE $s = 1$
  \STATE  \# Select $k$  points $\{x\}_{l=1}^{k} \subset X$
  \STATE $X^{(s)} = X \setminus \{x\}_{l=1}^{k} $
  \STATE \# Learn the first expert, $m_{s}$
  \STATE  \# Initialize $w_{s}$, weights of $m_{s}$
  \FOR{$j$ in $\{1,...,k\}$}
    \STATE \# Take gradient steps to minimize MoE loss:
    \STATE $w_{s}^{j} = w_ {s}^{j-1} - \nu \nabla E^{j}_{acc}$  
  \ENDFOR
  \STATE $\lambda_{1}=0, \lambda_{2}=0$, and  $\lambda_{3}=0$
  \WHILE{$X^{(s)}$ is not empty}
      \STATE s+=1
        \STATE $ \lambda_{1} = \lambda_{1} + \Delta \lambda_{1}, \lambda_{2} = \lambda_{2}+ \Delta \lambda_{2},\lambda_{3} = \lambda_{3}+\Delta\lambda_{3}$
      \STATE  \# Select $k$  points $\{x\}_{l=1}^{k} \subset X^{(s)}$
     \STATE $X^{(s)} = X^{(s)} \setminus \{x\}_{l=1}^{k} $
     \STATE \# Learn the subsequent expert, $m_{s}$
  \STATE Initialize $w_{s}$
  \FOR{$j$ in $\{1,...,s * k\}$}
    \STATE \# Take gradient steps to minimize Equation~\ref{moefinal}
    \FOR{$l$ in $\{1,..,s\}$}
    \STATE $w_{s}^{j} = w_ {s}^{j-1} - \nu \nabla E^{j}_{acc} - \lambda_{1}\nabla E_{SPD} - \lambda_{2} \nabla E_{AOD}^j - \lambda_{3}\nabla E_{Burden}^j $  
    \ENDFOR
  \ENDFOR
    \ENDWHILE
\end{algorithmic}

 \end{algorithm}
 
The overall loss for case $j$ is then given by: 
 
 \begin{equation}
\begin{aligned}
     E_{MOE}^j = E_{acc}^j + \lambda_{1}E_{SPD}^j + \lambda_{2}E_{AOD}^j + \lambda_{3}E_{Burden}^j 
\end{aligned}
\label{moefinal}
 \end{equation}

 \subsection{Data Drift and the FEAMOE algorithm}

Data Drift means that the statistical properties of the data, embodied in the underlying joint distribution of independent and dependent variables, can change over time, often in unforeseen ways. The change could be in the class priors, the class conditional distributions (concept drift), in the distribution of the independent variables etc. Drift can cause the model to become less accurate as time passes. However, drift can also cause other properties associated with the model to change, such as fairness. We develop an algorithm that can handle drift with respect to both accuracy and fairness.

Consider an online learning setup where input data points are observed sequentially. The algorithm to learn FEAMOE (Algorithm \ref{alg:algorithmmoe}) is as follows: start with a single model. Begin to train with data points (using stochastic gradient descent) and train the current model for a certain number of data points $k$ using only $E_{acc}$ (equation 2).
After $k$ points, introduce a new logistic regression model and train the mixture of experts with a softmax gating function using the loss in Equation \ref{moefinal}. Simultaneously, introduce the fairness penalties. Then, continue training for the next $k$ points, and then add another expert. As more experts are added, gradually increase the hyperparameters ($\lambda$' values) associated with the three fairness losses. This process is continued until all the available data is seen.

The motivation behind this training scheme is two-fold: in beginning with the accuracy penalty for the first expert, we ensure that the fairness measures do not interfere with training an accurate classifier, since high weights on the fairness terms would result in a less accurate classifier (as shown in experiments). Then, we slowly increase the weights on the fairness penalties with the goal that for individuals that are classified unfairly with respect to these group fairness measures, another expert takes over in this data regime to train for these individuals over time. This is because the mixture of experts framework allows some or all of the experts to learn on different regions of the data. Secondly, the algorithm allows us to account for drift, both with respect to the accuracy of the classifier and the fairness, since our framework allows for fairness constraints. If there is a change in the statistical properties of the data that impacts any of the loss terms, the mixture of experts adapts to this change over time through the addition of experts.

\subsection{Fast Shapley value explanations}

A prominent class of feature attribution methods is based on Shapley values from cooperative game theory \citep{shapley52}. Details about Shapley value explanations can be found in \citep{shap}, \citep{sundararajan2017axiomatic}, and \citep{aas2019explaining}. While computing Shapley values for a linear model is fast, doing so for non-linear models with methods like KernelShap \citep{lundberg2017unified} requires approximations and methods that cause the overall computation to become slow \citep{molnar2019,aas2019explaining}. Another method, TreeShap, \citep{lundberg2018consistent} works only for tree models. Though the mixture of experts model proposed is non-linear, as the individual experts are linear, the theorem below shows how to compute them for the whole model quickly and efficiently. 

Consider a mixture of experts model with $m$ experts. Let $\phi_{j}(m(\textbf{x}$)) be the Shapley value associated with expert $m$ for feature $j$ for an input instance $\textbf{x}$.

\begin{theorem}
For a mixture of experts model, the Shapley value for a given instance $\textbf{x}$ and feature $j$ for the model prediction is given by:
\begin{equation}
    \phi_j(y(\textbf{x})) = \sum_{i=1}^mg_i(\textbf{x})\phi_{j}(m(\textbf{x}))
\end{equation}
\label{thm:shap}
\end{theorem}





The proof is provided in the appendix. This result shows that the Shapley value for a given feature and input instance for the mixture of linear experts is a linear combination of the Shapley values of the feature and that input instance from every expert, weighted by the gating network's assigned weights for that input. 
This means that so long as the Shap values for individual models can be quickly computed (as is the case for linear/logistic regression, decision trees, XGBoost), the FEAMOE system-level Shap computation is also very quick.
In this paper, we illustrate FEAMOE using logistic regression experts, so this desirable property holds, even though the mixture model is able to construct non-linear models of arbitrary complexity by including as many logistic regression-based experts as needed. 

\section{Experiments and Results}

\begin{table*}[t!]
\caption{Names of the models compared in figure \ref{fig:UCI Adult comparison}, based on the class of models (Mixture of Experts (MOE) or Neural Network (NN)) and type of fairness constraints (None, SPD, AOD, Burden, or all). Experiments on models marked x are in the appendix}
\centering
\begin{tabular}{cccccc} 
\hline
Type &   None & SPD & AOD & Burden & All\\
\hline
MOE    &  MOE & FEAMOE1 & FEAMOE2 & FEAMOE3 & FEAMOE \\
NN     &  NN & x & x & x& FairNN \\
\hline
\end{tabular}
\label{Tablenamemodel}

\end{table*}

\begin{figure*}
     \centering
     \begin{subfigure}[b]{0.463\textwidth}
         \centering
         \includegraphics[width=\textwidth]{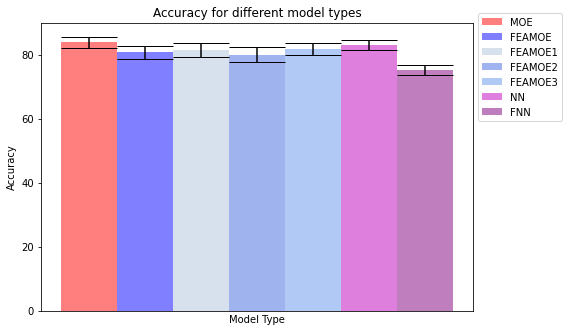}
         \caption{}
         \label{fig:3a}
     \end{subfigure}
     \hfill
     \begin{subfigure}[b]{0.463\textwidth}
         \centering
         \includegraphics[width=\textwidth]{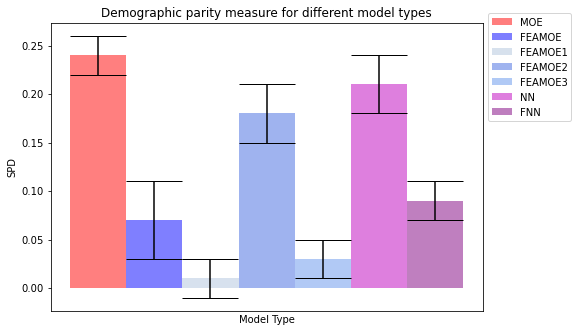}
         \caption{}
         \label{fig:3b}
     \end{subfigure}
     \hfill
     \begin{subfigure}[b]{0.463\textwidth}
         \centering
         \includegraphics[width=\textwidth]{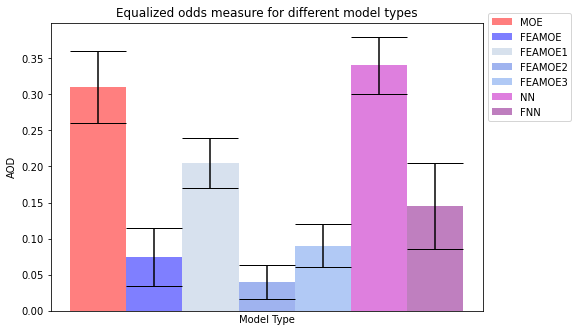}
         \caption{}
         \label{fig:3c}
     \end{subfigure}
          \hfill
     \begin{subfigure}[b]{0.463\textwidth}
         \centering
         \includegraphics[width=\textwidth]{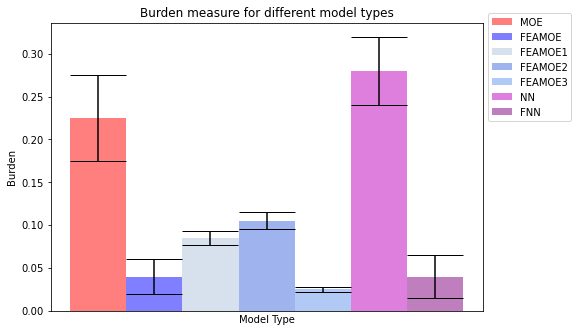}
         \caption{}
         \label{fig:3d}
     \end{subfigure}
        \caption{Results for the UCI Adult Dataset on different fairness constraints being incorporated. Details on model names are provided in Table \ref{Tablenamemodel}.  A higher accuracy is better and a lower bias is better}
        \label{fig:UCI Adult comparison}
\end{figure*}

Experiments are performed using the mixture of experts with logistic regression experts and a softmax gating function. We implement the logistic regression models using scikit learn with default parameters. We show that using logistic regression experts within the MOE produces accuracies similar to using appropriately sized neural networks while allowing for the generation of faster explanations. All neural networks are multilayer perceptrons. There are two sets of experiments, highlighting different aspects of FEAMOE:\\
(a) {\bf Fairness Study}. We use two classification datasets that exhibit bias and, despite having some known issues \citep{ding2021retiring}, are very well studied in the fairness community: UCI Adult \citep{kohavi1996scaling} and COMPAS \citep{ProPublica}.
Note that these datasets are not inherently streaming; so they are being used only for fairness studies rather than for handling of drifts. Gender is considered as the protected attribute for UCI Adult. 
A two layer multilayer perceptron with 30 hidden
units in each layer was trained for the UCI Adult dataset. Additional or larger hidden layers, or ensemble methods such as xgboost do not provide extra benefit for these two tabular datasets, and hence are omitted for comparison purposes. Experiments on the COMPAS dataset are in the appendix. \\
(b) {\bf Drift Study}. The large HMDA (Home Mortgage Disclosure Act) dataset \citep{HMDA} reflects data from multiple years, with the underlying data statistics varying considerably over the years, so it is suitable for drift studies. Gender is the protected attribute.
A five layer multilayer perceptron with 50 hidden units in each layer is trained for the HMDA dataset as the baseline neural network. Experiments on a synthetic streaming version of the UCI Adult dataset are in the appendix.
Data sequencing for training is described later; for now we mention that batch techniques (e.g. standard tree based models), including popular ensemble techniques, that make multiple passes over data sampled from the entire time span, cannot be deployed as we are studying effects of drift over time.

First, we use UCI Adult to demonstrate the effects of incorporating the proposed fairness constraints on the mixture of experts models. Similar results for the COMPAS dataset are provided in the appendix. We then show that the HMDA dataset demonstrates drift with respect to both fairness and accuracy, and that FEAMOE can adapt to such drifts. Comparisons are made to neural networks (both with and without fairness constraints), which is the state-of-the-art model class for accuracy-based performance across these datasets in all experiments. Experiments on faster Shapley value explanations are in the appendix.

\subsection{Fairness Constraints}

Experiments are performed on UCI Adult in seven different regimes based on model types and fairness constraints. Details of these regimes are in Table \ref{Tablenamemodel}. We use our training algorithm such that experts are added every 4000 data points for the UCI Adult dataset. Hyperparameters associated with the fairness constraints are incremented in levels of 0.02 per expert for the UCI Adult dataset. The parameters are found using grid search and vary based on dataset size and extent of prevalent bias (details in appendix). The results are averaged across five runs. We report the accuracy and the absolute value of the three fairness measures (for consistency in interpreting results across fairness measures). We provide comparisons to other methods for bias reduction \citep{calmon2017optimized,sharma2020fair,agarwal2018reductions} in the appendix.


The results are shown in Figure \ref{fig:UCI Adult comparison}. The accuracy across different model types remains similar for the UCI Adult dataset, but using just a neural network with fairness constraints works poorly, as shown in Figure \ref{fig:UCI Adult comparison}a. As seen in \ref{fig:UCI Adult comparison}b,c,d, the fairness measures also work well even in isolation from each other That is, in trying to improve based on just one measure, the other measures also improve. In this regard, the burden-based fairness measure (FEAMOE3) has the best effect; just using burden-based fairness alone helps significantly improve the other fairness measures while maintaining reasonable accuracy. This behavior agrees with the observations in \citep{sharma2020fair}.  The fair neural network (FairNN) performs worse for demographic parity and equalized odds compared to FEAMOE. We hypothesize that this happens because our learning process slowly induces fairness with every expert, as opposed to training the whole architecture in one go. Overall, FEAMOE significantly reduces all three forms of biases while maintaining accuracy.

\begin{figure*}
     \centering
     \begin{subfigure}[b]{0.47\textwidth}
         \centering
         \includegraphics[width=\textwidth]{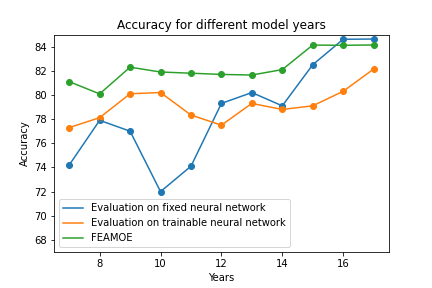}
         \caption{}
         \label{fig:6a}
     \end{subfigure}
     \hfill
     \begin{subfigure}[b]{0.47\textwidth}
         \centering
         \includegraphics[width=\textwidth]{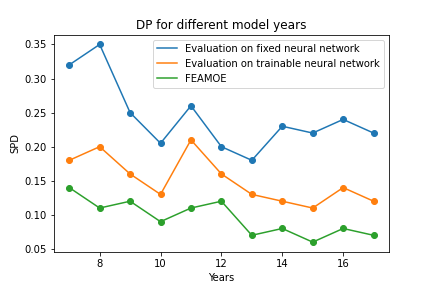}
         \caption{}
         \label{fig:6b}
     \end{subfigure}
     \hfill
     \begin{subfigure}[b]{0.47\textwidth}
         \centering
         \includegraphics[width=\textwidth]{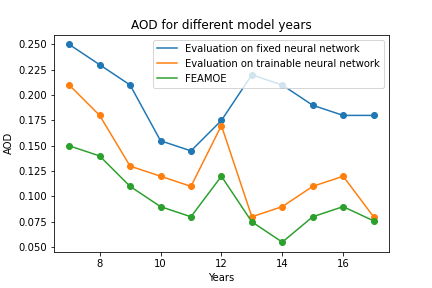}
         \caption{}
         \label{fig:6c}
     \end{subfigure}
          \hfill
     \begin{subfigure}[b]{0.47\textwidth}
         \centering
         \includegraphics[width=\textwidth]{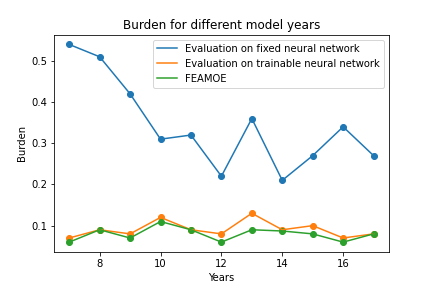}
         \caption{}
         \label{fig:6d}
     \end{subfigure}
        \caption{Comparison of Drift Handling on the HMDA dataset. Lower bias values are better. Results with error bars (excluded here for readability) are provided in the appendix. 
1) Blue: baseline neural networks (fixed neural network) trained without fairness constraints on a previous year (20XX, indicated by x-axis; 2016 and 2017 are the "future" years) and not updated with new data; 2) Orange: (fair and) trainable neural network: Neural networks with fairness constraints incorporated; also updated with streaming data from the "future" years and 3) Green: FEAMOE, also update with a single pass on streaming data from the "future" years. Note that several popular models (including ensembles such as XGBoost) are not being considered as by default they will need to make multiple passes over the dataset and are really not designed for streaming (single-look) applications. 
        }
        \label{fig:hmdamitigation}
\end{figure*}
\subsection{Real-World Drift: The HMDA dataset}
We first demonstrate that the HMDA dataset exhibits drift across years, and then show FEAMOE's effectiveness in handling it. The HMDA dataset has millions of records of individuals spanning several years. It contains consumer characteristics; the target variable indicates whether a consumer received a mortgage. While this dataset considered as a whole has been previously shown to exhibit bias, there is no investigation into how such bias varies across the years. First, to quantify drift in this dataset in both fairness and accuracy, we trained one neural network per year from 2007 to 2017, each on 100,000 random samples in that year, and tested each of these networks on data from the years 2016-2017 (Once trained, these models, which we call fixed neural networks, cannot be updated). The results are shown in Figure \ref{fig:hmdamitigation} by the blue points. In general, the farther the training data is away from the test year the more the accuracy and fairness measures degrade (i.e., accuracy decreases and fairness differences increase). Also training a single model on a dataset of the same size but sampled uniformly over all the previous years does not help either as the data is non-stationary. To the best of our knowledge, this is so far the most detailed study on fairness drift over time, made on an openly accessible, real-world, large-scale dataset. 

We now study how fairness aware neural networks with online updates compare with FEAMOE in their ability to handle drift. Note that for our setting, we cannot use certain models such as popular ensemble approaches (XGBoost etc), that need multiple passes through new data in batch mode after the initial model is built and deployed. 
For FEAMOE, we train models for each year separately and consider each of those as one expert. Then, we add on experts based on a single pass on new data (from the 2016-2017 years).  
We compare this to neural networks (with fairness constraints) that are also trained on each of the past years in batch mode, and then updated online with a single pass on the new data. The FEAMOE results are shown in Figure \ref{fig:hmdamitigation} by the green points, and the trainable neural networks results by orange points.   FEAMOE is noticeably better at maintaining good accuracy and keeping lower bias across all bias metrics, pretty much irrespective on how old the original model was, even when compared to adaptive neural networks. We believe that the loosely coupled architecture and adaptive model complexity is key to FEAMOE's success in handling drift. It is also worth noting that using the FEAMOE architecture provides for much faster Shapley value explanations compared to the trainable neural network (as shown in experiments in the supplementary material).

\subsection{Additional Experiments and Findings}

We performed several more experiments to validate the effectiveness of using FEAMOE. We summarize the findings here; further details are in the appendix. Irrespective of dataset size, using neural networks as experts instead of logistic regression experts in the mixture had negligible effects on the accuracy. We also tried variants of the proposed learning method, such as introducing experts based on individual fairness-based performance saturation instead of adding experts at (predetermined) regular intervals, achieving good performance. Having more experts with a smaller increment on fairness constraints was also considered and leads to similar performance as having fewer experts with a tuned increment on the fairness hyperparameters across datasets. Further details on hyperparameters can also be found in the appendix. Experiments on fast Shapley value explanations are in the appendix. Also, experiments on the COMPAS dataset for the fairness study, and on the synthetic streaming version of the UCI Adult dataset for the drift study can be found in the appendix. 

\section{Conclusion and Future Work}
We propose FEAMOE: a novel mixture of experts architecture and learning framework that can better maintain the fairness of a model in the face of data drift. We show how three fairness constraints can be incorporated into this framework. These constraints are illustrative, as one can use alternative constraints instead. We prove that by using this mixture of experts, Shapley value explanations can be computed efficiently even though the overall model is non-linear. Experiments are performed on three datasets to demonstrate the various properties and effectiveness of FEAMOE. 
In particular, we identified a large-scale, real-world dataset that induces drift with respect to fairness over time in non-adaptive models, and show that our framework can adequately address this challenge. 
Given the usefulness of the FEAMOE formulation, we would now like to extend it to incorporate other potential forms of drift, such as those that cause changes in adversarial robustness.

\bibliographystyle{ACM-Reference-Format}
\bibliography{fairn_aies1.bib}
\newpage
\appendix
\section{Supplementary Material}

We present the supplementary material for FEAMOE. The key items provided are:
\begin{itemize}
    \item Proof of Theorem 1
    \item Experiments on models marked x in Table 1 of the paper
    \item Additional details on experiment setups and finding hyperparameters to reproduce experiments
    \item COMPAS dataset experiments
    \item Experiments on synthetic drift using the UCI Adult dataset
    \item HMDA experiments with error bars
    \item Experiments on fast Shapley value explanations
    \item Experiments on comparison to existing fairness mitigation strategies
    \item Experiments on comparison of accuracy of using neural network experts to logistic regression experts
    \item Individual fairness based experiments
    \item Experiments on using more experts with less increments on the fairness constraints
    \item Example Shapley value explanations
    
\end{itemize}

\subsection{Proof of Theorem 1}

Consider a mixture of experts model with $m$ experts. Let $\phi_{j}(m(\textbf{x}$)) be the Shapley value associated with expert $m$ for feature $j$ for an input instance $\textbf{x}$.

For a mixture of experts model, the Shapley value for a given instance $\textbf{x}$ and feature $j$ for the model prediction is given by:
\begin{equation}
    \phi_j(y(\textbf{x})) = \sum_{i=1}^mg_i(\textbf{x})\phi_{j}(m(\textbf{x}))
\end{equation}
\label{thm:shap}

For any given input $\textbf{x}$, the prediction for the mixture of experts is simply a weighted combination of the predictions from individual experts. Using the linearity property of Shapley values \citep{shapley1953stochastic}:

\begin{equation}
\begin{aligned}
    \phi_j(y(\textbf{x})) & = \phi_j(g_1(\textbf{x})y_1(\textbf{x}) + g_2(\textbf{x})y_2(\textbf{x})+ ...+g_m(\textbf{x})y_m(\textbf{x})) \\
    & = \phi_j(g_1(\textbf{x})y_1(\textbf{x})) + \phi_j(g_2(\textbf{x})y_2(\textbf{x})) + ... + \phi_j(g_m(\textbf{x})y_m(\textbf{x})) \\
    &= g_1(\textbf{x})\phi_j(y_1(\textbf{x})) + g_2(\textbf{x})\phi_j(y_2(\textbf{x})) + ... + g_m(\textbf{x})\phi_j(y_m(\textbf{x})) \\
    & = \sum_{i=1}^mg_i(\textbf{x})\phi_{j}(m(\textbf{x}))
\end{aligned}
\end{equation}

\subsection{Experiments on Neural Network Models from Table 1}

\begin{figure*}[h!]
  \centering 
  \includegraphics[scale=.4]{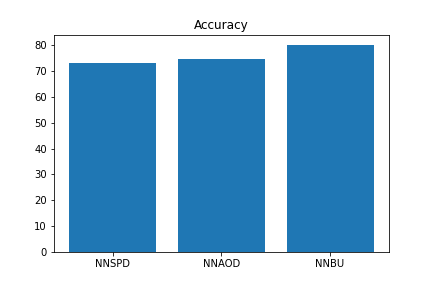}
  \includegraphics[scale=.4]{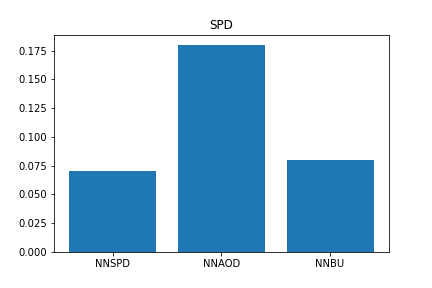} 
  \includegraphics[scale=.4]{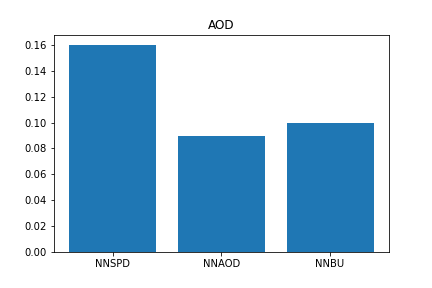}
  \includegraphics[scale=.4]{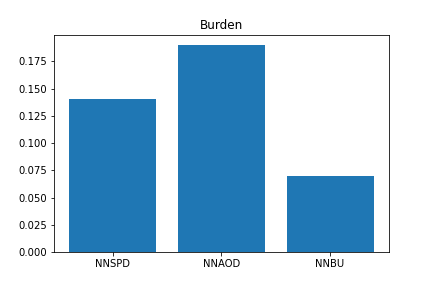}

  \caption{Results for the UCI Adult Dataset on different fairness constraints being incorporated for neural network models. A higher accuracy is better and a lower fairness value is better}
  \label{fig:UCIAdultNN}
\end{figure*}

Results for fairness constraints incorporated independently in neural network models are shown in Figure \ref{fig:UCIAdultNN}. As we can see, the fairness constraints work even with neural networks, but when compared to results provided in the main draft, they do not work as well as with the equivalent FEAMOE models (FEAMOE1, FEAMOE2, and FEAMOE3 respectively).

\subsection{Additional Details on Experiments}

\begin{table*}[t!]
\caption{Results on varying the number of data points $k$ after which an expert is added for the UCI Adult dataset. Here the increase in the weights of fairness penalties per expert added is fixed at 0.02.}
\centering
\begin{tabular}{cccccc} 
\hline
Performance &   $k=100$ & $k=1000$ & $k=4000$ & $k=10000$ & $k=41018$ \\
\hline
Accuracy    &  46.34 & 74.77 & 79.20 & 83.1 & 77.54  \\
SPD      &  0.002 & 0.007 & 0.06 & 0.14 & 0.26 \\
AOD    & 0.002 &  0.03 &   0.07    &    0.24   &  0.31    \\
Burden    & 0.01 &  0.02 &    0.038  & 0.25 &  0.28  \\
\hline
\end{tabular}
 \label{Table1}

\end{table*}

To study the effect of changing the number of experts on performance, we vary the number of data points $k$ after which an expert is added for the UCI Adult dataset for a fixed increment in the fairness associated hyperparameters, and study the effect. For example, to obtain one expert (i.e. simply a logistic regression model), $k=n$, where $n$ is the size of the dataset.

The results are shown in Table \ref{Table1}. When the number of experts is large ($k$=100) with a fixed increment of 0.02 for each fairness measure, the fairness weights start dominating significantly with just a few experts, causing the model to become inaccurate. With fewer experts for the same increment of each fairness measure weight, the accuracy starts to improve. For $k=4000$, which represents 11 experts, the model learned is reasonably accurate and fairer. Decreasing the number of experts further results in a more accurate classifier to a point, but fairness measures suffer since the weight associated with them is very low. Finally, just having one logistic regression model with no constraint on fairness ($k$ = 41018, which is the total number of data points) results in a poor model in performance, since this model is just a linear model. This highlights the importance of tuning the number of experts to find a balance between fairness and accuracy for this framework.

Experiments are performed using the mixture of experts with logistic regression experts and a softmax gating function. We consider three datasets that exhibit bias: UCI Adult \cite{kohavi1996scaling}, COMPAS \cite{ProPublica}, and the HMDA (Home Mortgage Disclosure Act) datasets \cite{HMDA}. The UCI Adult dataset pertains to the financial domain, where the goal is to predict a class of income ($>$ 50k is the positive outcome). Gender is considered as the protected attribute. The COMPAS dataset is a classification dataset where the prediction variable is the likelihood to re-offend. Race is considered as the protected attribute. The HMDA dataset is used to predict if an individual receives a mortgage or not. Race is considered as the protected attribute. The dataset exists across several years and has a huge number of data entries in every year. We use the UCI Adult and COMPAS dataset to demonstrate the effects of incorporating the proposed fairness constraints on the mixture of experts models. The UCI Adult dataset is used to show the ability to handle drift when drift with respect to fairness is synthetically induced. The HMDA dataset is used to show a real world application of our approach towards handling concept drift. Experiments were done on a standard laptop with an Intel i5 processor and no GPU's were used. 

The neural networks used to compare with FEAMOE were 2 hidden layer neural networks with 30 neurons in each layer for the UCI Adult and COMPAS datasets, and 5 hidden layer neural networks with 50 neurons in each layer for the HMDA dataset. The number of experts in FEAMOE depends on the number of data points considered after which an expert is added. The number of data points, along with the hyperparameters for every fairness constraint can be found using grid search. Through extensive experimentation, we found that the number of points after which experts are added should be around one-tenth of the expected data size if data is sequential or dataset size if the whole dataset is available. The fairness constraints-based hyperparameters (increments) should not be beyond a factor of 0.1 otherwise the models have poor accuracy. Also, since the burden constraint works reasonably well to also help with SPD and AOD, only that constraint can be considered to avoid tuning for four different hyperparameters. We provide code for a basic notebook implementing mixture of experts with fairness constraints. All experiments are simple variations of this notebook.

\subsection{COMPAS dataset experiments}

\begin{figure*}
     \centering
     \begin{subfigure}[b]{0.45\textwidth}
         \centering
         \includegraphics[width=\textwidth]{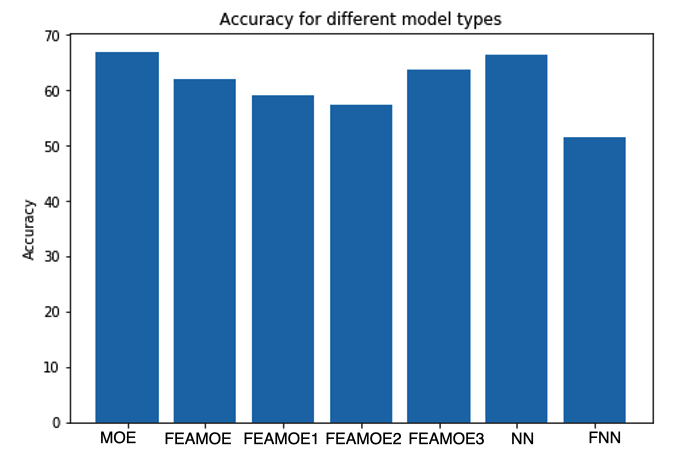}
         \caption{}
         \label{fig:4a}
     \end{subfigure}
     \hfill
     \begin{subfigure}[b]{0.47\textwidth}
         \centering
         \includegraphics[width=\textwidth]{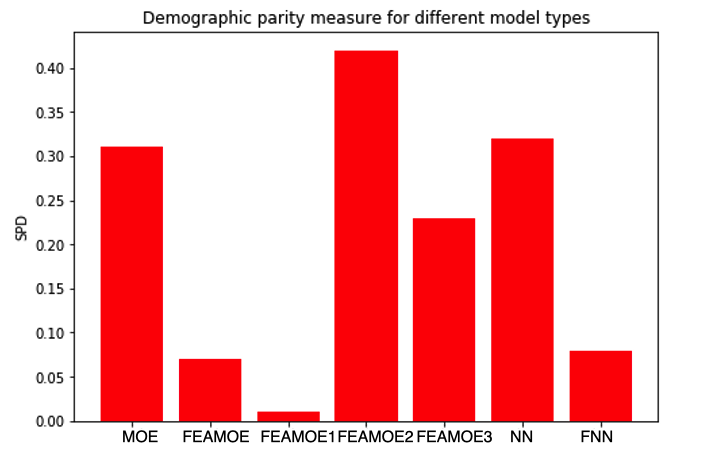}
         \caption{}
         \label{fig:4b}
     \end{subfigure}
     \hfill
     \begin{subfigure}[b]{0.45\textwidth}
         \centering
         \includegraphics[width=\textwidth]{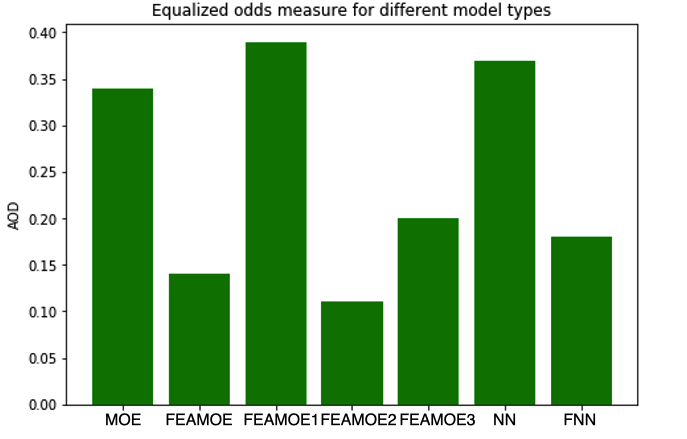}
         \caption{}
         \label{fig:4c}
     \end{subfigure}
          \hfill
     \begin{subfigure}[b]{0.47\textwidth}
         \centering
         \includegraphics[width=\textwidth]{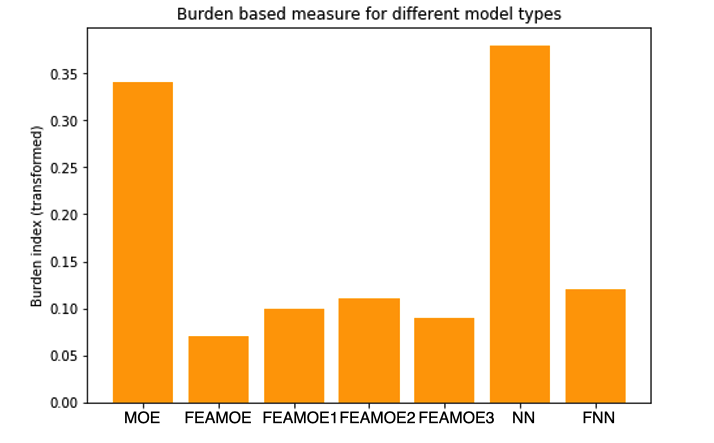}
         \caption{}
         \label{fig:4d}
     \end{subfigure}
        \caption{Results for the COMPAS dataset on different fairness constraints being incorporated. Details on model names are provided in Table \ref{Tablenamemodel}. A higher accuracy is better and a lower fairness measure value is better}
        \label{fig:COMPAS Comparison}
\end{figure*}

Compared to the UCI Adult dataset, the COMPAS dataset (results in Figure \ref{fig:COMPAS Comparison}) has more of an accuracy drop with the inclusion of fairness constraints (Figure \ref{fig:COMPAS Comparison}a), due to an inherent fairness accuracy trade-off, as has been shown before \citep{calmon2017optimized, sharma2020data}. This can be attributed to this dataset having a prejudice-based bias or label bias as opposed to the UCI Adult dataset which has a sampling-based bias \citep{sharma2020data}. It can also be observed that for this dataset, there is an inherent trade-off between demographic parity and equalized odds (Figure \ref{fig:COMPAS Comparison}b,c). In trying to improve just one, the other worsens (FEAMOE1 and FEAMOE2). The burden-based constraint performs similarly to the effect it had for the UCI Adult dataset (Figure \ref{fig:COMPAS Comparison}d). Since burden-based fairness depends on the distance of points to the boundary, a classifier may be able to maintain its accuracy while changing the distance to points to reduce the recourse gap. This effect can be seen through the COMPAS dataset: the accuracy does not drop significantly when only the burden constraint is used. However, including all three constraints (FEAMOE) produces a model that is fairer than standard models and reasonably accurate, thereby highlighting the importance of the inclusion of all three constraints in the objective. 

\subsection{Experiments on synthetic drift using the UCI Adult dataset}

\begin{figure*}
     \centering
     \begin{subfigure}[b]{0.47\textwidth}
         \centering
         \includegraphics[width=\textwidth]{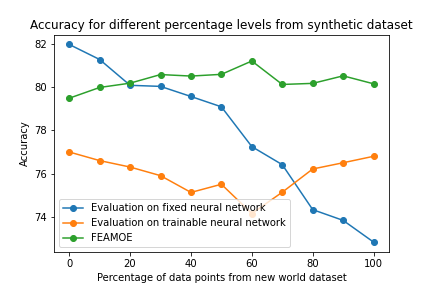}
         \caption{}
         \label{fig:5a}
     \end{subfigure}
     \hfill
     \begin{subfigure}[b]{0.47\textwidth}
         \centering
         \includegraphics[width=\textwidth]{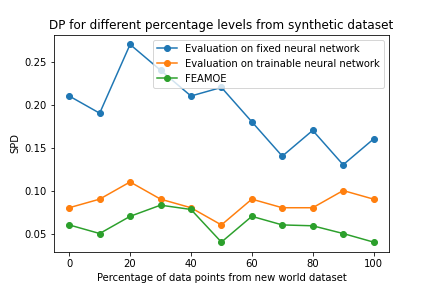}
         \caption{}
         \label{fig:5b}
     \end{subfigure}
     \hfill
     \begin{subfigure}[b]{0.47\textwidth}
         \centering
         \includegraphics[width=\textwidth]{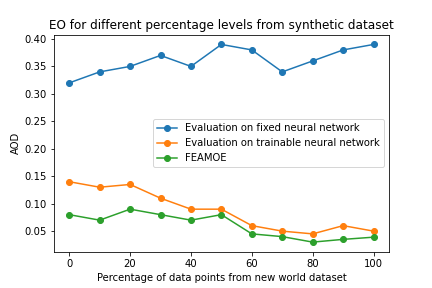}
         \caption{}
         \label{fig:5c}
     \end{subfigure}
          \hfill
     \begin{subfigure}[b]{0.47\textwidth}
         \centering
         \includegraphics[width=\textwidth]{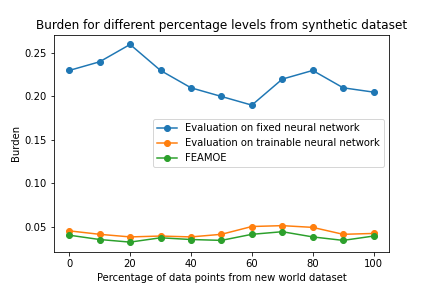}
         \caption{}
         \label{fig:5d}
     \end{subfigure}
        \caption{Results on the ability to handle drifts of different magnitudes. When a lower percentage of data points are considered from the new world dataset (which is the low level of augmentation from \citep{sharma2020data}), less drift occurs. However, with an increase in data points from the sorted new world dataset (sorted in decreasing order of how realistic points are to the original dataset), there is more drift, both with respect to accuracy and fairness. Higher accuracy is better and lower fairness value is better. Fixed neural network is trained on the original dataset without fairness constraints and tested on the new world dataset subsets. Trainable neural network is trained on the original dataset with fairness constraints, and is retrained in online mode with the subsets from the new world dataset}
        \label{fig:Adultacroosperc}
\end{figure*}


To test FEAMOE on drift that could occur due to a change in the characteristics of groups within a dataset, we perform experiments on datasets synthetically generated based on the UCI Adult dataset. We perform experiments on synthetically induced drift to show that our method can handle drifts of different magnitudes. 

To synthetically induce drift, we generate a dataset using the UCI Adult dataset where the binary protected attribute gender is flipped, as in \citep{sharma2020data}. We call this the new world dataset. This new dataset has more females belonging that receive an income of greater than 50k per year. To assess the ability to handle drifts of different magnitudes, we consider sequentially chosen subsets from the new world dataset where the data is sorted in order of realism as in \citep{sharma2020data}. In \citep{sharma2020data}, realism of a data point is determined by the distance between the data point in the new dataset with cluster centers (found using k-means) from the original dataset. For example, we first consider only the first $10\%$ of the most realistic new world data, which is the $10\%$ augmentation level in \citep{sharma2020data}. The most realistic subset would induce the least drift (slow drift), and as the distance of the synthetic subsets to the original dataset increases, the drift is more pronounced. As a baseline, we train a neural network (without any fairness constraint) on the original UCI Adult dataset. We call this the fixed neural network. Then, we test this network on data from different subsets of the new world data. In testing this model, we go from the most to least realistic subsets (i.e. less drift to more drift) and track the accuracy and the three fairness measures introduced earlier. We compare this model to training our adaptive mixture of logistic regression experts framework on these subsets. We also compare to training the original neural network in online mode with fairness constraints on these subsets i.e. adapting a neural network with fairness constraints. We call this the trainable neural network.

The results are shown in Figure \ref{fig:Adultacroosperc}. Both the accuracy and fairness measures remain consistently poor for the baseline neural network (fixed neural network), and they degrade as subsets from less realistic portions of the new world data are evaluated, demonstrating that drift with respect to both accuracy and fairness is possible when test data has drifted from the training data. We need frameworks that can adapt to this. To implement FEAMOE, we train a logistic regression model with the original UCI Adult dataset, and then add experts for the new subsets based on Algorithm 1. FEAMOE can account for different levels of drift. The accuracy of our model remains fairly constant, and the bias is significantly lower compared to the baseline model. Our method is also able to adapt better to maintain accuracy and all fairness measures, compared to a neural network that can be trained with the new data points and with fairness constraints (i.e. the trainable neural network).

\subsection{HMDA experiments with error bars}

We first demonstrate that the HMDA dataset exhibits drift across years, and then show FEAMOE's effectiveness in handling it. The HMDA dataset has millions of records of individuals spanning several years. It contains consumer characteristics; the target variable indicates whether a consumer received a mortgage. While this dataset considered as a whole has been previously shown to exhibit bias, there is no investigation into how such bias varies across the years. First, to quantify drift in this dataset in both fairness and accuracy, we trained one neural network per year from 2007 to 2017, each on 100,000 random samples in that year, and tested each of these networks on data from the years 2016-2017 (Once trained, these models, which we call fixed neural networks, cannot be updated). The results are shown in Figure \ref{fig:hmdamitigation} by the blue points. In general, the farther the training data is away from the test year the more the accuracy and fairness measures degrade (i.e., accuracy decreases and fairness differences increase). Also training a single model on a dataset of the same size but sampled uniformly over all the previous years does not help either as the data is non-stationary. To the best of our knowledge, this is so far the most detailed study on fairness drift over time, made on an openly accessible, real-world, large-scale dataset. 

We now study how fairness aware neural networks with online updates compare with FEAMOE in their ability to handle drift. Note that for our setting, we cannot use certain models such as popular ensemble approaches (XGBoost etc), that need multiple passes through new data in batch mode after the initial model is built and deployed. 
For FEAMOE, we train models for each year separately and consider each of those as one expert. Then, we add on experts based on a single pass on new data (from the 2016-2017 years).  
We compare this to neural networks (with fairness constraints) that are also trained on each of the past years in batch mode, and then updated online with a single pass on the new data. The FEAMOE results are shown in Figure \ref{fig:hmdamitigation} by the green points, and the trainable neural networks results by orange points.   FEAMOE is noticeably better at maintaining good accuracy and keeping lower bias across all bias metrics, pretty much irrespective on how old the original model was, even when compared to adaptive neural networks. We believe that the loosely coupled architecture and adaptive model complexity is key to FEAMOE's success in handling drift. It is also worth noting that using the FEAMOE architecture provides for much faster Shapley value explanations compared to the trainable neural network (as shown in experiments later).

\begin{figure*}
     \centering
     \begin{subfigure}[b]{0.47\textwidth}
         \centering
         \includegraphics[width=\textwidth]{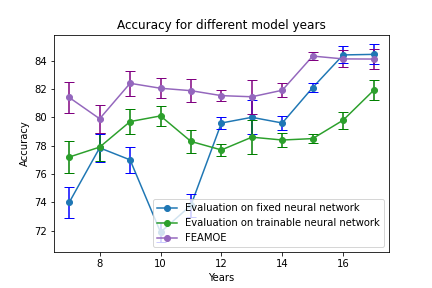}
         \caption{}
         \label{fig:6a}
     \end{subfigure}
     \hfill
     \begin{subfigure}[b]{0.47\textwidth}
         \centering
         \includegraphics[width=\textwidth]{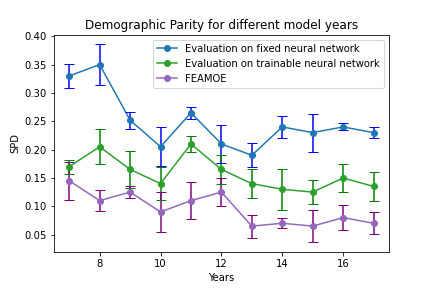}
         \caption{}
         \label{fig:6b}
     \end{subfigure}
     \hfill
     \begin{subfigure}[b]{0.47\textwidth}
         \centering
         \includegraphics[width=\textwidth]{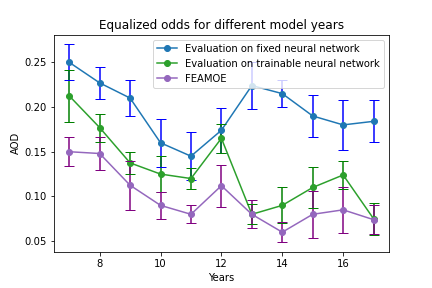}
         \caption{}
         \label{fig:6c}
     \end{subfigure}
          \hfill
     \begin{subfigure}[b]{0.47\textwidth}
         \centering
         \includegraphics[width=\textwidth]{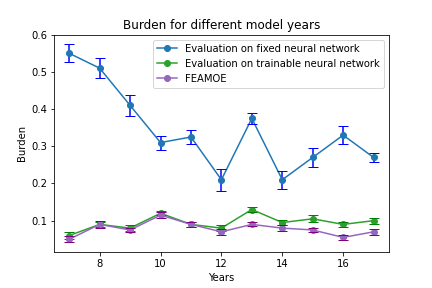}
         \caption{}
         \label{fig:6d}
     \end{subfigure}
        \caption{Comparison of Drift Handling on the HMDA dataset. Lower bias values are better. Results with error bars included.
1) Blue: baseline neural networks (fixed neural network) trained without fairness constraints on a previous year (20XX, indicated by x-axis; 2016 and 2017 are the "future" years) and not updated with new data; 2) Orange: (fair and) trainable neural network: Neural networks with fairness constraints incorporated; also updated with streaming data from the "future" years and 3) Green: FEAMOE, also update with a single pass on streaming data from the "future" years. Note that several popular models (including ensembles such as XGBoost) are not being considered as by default they will need to make multiple passes over the dataset and are really not designed for streaming (single-look) applications. 
        }
        \label{fig:hmdamitigation}
\end{figure*}

\subsection{Shapley Value Explanations}

\begin{table*}[t!]
\caption{Results on comparing KernelShap with using LinearShap for individual experts and combining them using Theorem 1 (FEAMOE) for 100 instances across three datasets. Two variants of KernelShap are considered, with varying the number of samples $n$ allowed to form a combination of features for computation. Time is measured in seconds}
\centering
\begin{tabular}{cccc} 
\hline
Dataset &   FEAMOE Time & KernelShap Time $n=500$ & KernelShap Time $n=2000$ \\
\hline
UCI Adult    &  34 & 379 & 941   \\
COMPAS      &  32 & 246 & 805  \\
HMDA    & 38 &  314 &  887    \\
\hline
\end{tabular}
\label{TableShap}

\end{table*}

As an added benefit, by using a mixture of experts with logistic regression experts, we can generate exact explanations in a computationally efficient manner. We generate Shap values for each independent expert in FEAMOE using LinearShap from the SHAP explainer \citep{lundberg2017unified} and then calculate the final Shapley value using Theorem 1. We compare FEAMOE's explanation computation time to KernelShap.  We do this for 100 instances across the three datasets. We compare the two methods based on the time taken across the 100 samples. Table \ref{TableShap} shows the results. As we can see, KernelShap takes significantly more time than using FEAMOE to compute the final Shapley values. When the allowed feature combinations are increased towards the computation of the KernelShap approximation (i.e. $n$ is increased to produce explanations closer to exact Shapley values), it takes even more time. Hence, our method generates explanations much faster and can scale better for large numbers of features and observations. This is particularly useful in a production setting that requires a huge number of explanations quickly.

\subsection{Comparison to other fairness mitigation methods}

We compare using the fairness constraints used in FEAMOE to existing state of the art bias mitigation methods for neural networks for the UCI Adult Dataset. The results are shown in Table \ref{tab:comparison}. As we can see, FEAMOE produces accurate and fair models across all bias measures. Other methods optimize on only one or two of the measures, but we still compare across all measures to validate that the constraints introduced in FEAMOE can sufficiently reduce bias and while maintaining accuracy. 

\begin{table*}[t!]
    \centering
    \begin{tabular}{ccccc}
    \hline
      Method & Accuracy & SPD & AOD & Burden \\
     \hline
     NN & 82.02 & 0.19 & 0.29 & 0.29 \\
     FEAMOE & 83.24 & 0.06 & 0.07 & 0.03\\
     FaiR-N-4 &  83.75 & 0.12 & 0.11 & 0.02  \\
  P.R.$^\dagger$ &  78.14 & 0.09 & 0.14 & 0.32 \\
     O.P.P.$^\dagger$ & 77.62 & 0.10 & 0.08 & 0.16 \\
     DataAug$^\dagger$ & 78.57 & 0.07 & 0.06 & 0.11 \\
     RedApp$^\dagger$ & 81.98 & 0.10 & 0.08 & 0.21 \\
     \hline
    \end{tabular}
    \caption{Comparison of models (averaged across five runs) trained on the Adult dataset. NN is a neural network trained without any fairness constraints, FaiR-N-4 is the model from \cite{sharma2020fair}, $^\dagger$: prejudice remover (P.R.) \cite{kamishima2012fairness}, optimized pre-processing (O.P.P.) \cite{calmon2017optimized}, DataAug \cite{sharma2020data}, and RedApp \cite{agarwal2018reductions}}    \label{tab:comparison}
\end{table*}

\subsection{Using neural network experts in FEAMOE}

Instead of using logistic regression experts, neural network experts can also be used, with the disadvantage of not being able to get fast explanations. However, we compare using neural network experts with using logistic regression experts to see if there is a difference in performance for the HMDA dataset for data from the year 2017. 

The results are shown in Table \ref{tab:NNLR}. The table shows how across accuracy and fairness constraints, neural network experts and logistic regression perform comparably. While the number of logistic regression experts needed is more than the number of neural network experts needed for similar performance, using logistic regression models comes with the advantage of providing explainability. 

\begin{table*}[h!]
\centering
\begin{tabular}{ccc} 
\hline
Performance &   NN & LR (FEAMOE) \\
\hline
Accuracy    &  84.21 & 84.08  \\
SPD      &  0.08 & 0.07  \\
AOD    & 0.078 &  0.081    \\
Burden    & 0.08 &  0.08  \\
Number of Experts & 8 & 13 \\
\hline
\end{tabular}
  \caption{Results on using neural network experts and compared to using logistic regression experts} \label{Table1}

\label{tab:NNLR}
\end{table*}

\subsection{Experiment on using individual fairness}

To use a performance saturation based expert growing mechanism instead of growing based on the number of data points, we tested for individual fairness and only added an expert when a certain number of data points $n$ did not satisfy individual fairness. Also, we add a constant penalty to the original mixture of experts accuracy loss such that it is 0 when the point is individually fair and is 0.4 (can be changed) when the point is not individually fair. We tested individual fairness by flipping the protected attribute and evaluating the model: if the flipped data point has the same prediction as the original data point, this input data point satisfies individual fairness, otherwise it does not. When $n$ points do not satisfy individual fairness, we add another expert. (Note that this is a naive method to check for individual fairness, however any other mechanism can be adopted to test for individual fairness). 

The results on accuracy and individual fairness for FEAMOE for the COMPAS dataset are shown in Table \ref{Table2}, where individual fairness is measured as the fraction of points that do not satisfy individual fairness in the dataset and compared to a neural network. As we can see, FEAMOE is still reasonably accurate, while ensuring that the classifier is more individually fair. 

\begin{table*}[h!]
\centering
\begin{tabular}{ccc} 
\hline
Performance &   NN & FEAMOE \\
\hline
Accuracy    &  66.18 & 64.13  \\
Individual Fairness & 0.38 & 0.09\\
\hline
\end{tabular}
  \caption{Results on using individual fairness in FEAMOE} \label{Table2}

\label{tab:NNLR}
\end{table*}

\subsection{Including more experts}

We show that using more experts (i.e. adding each expert after less number of data points) but with smaller changes in fairness hyperparameters (increments to $\lambda1$, $\lambda2$, and $\lambda 3$). For the COMPAS dataset, we add experts after every 40 data points with a fairness increment of 0.002 with every expert added. The results are in Table \ref{Table1}. As we can see, the results are comparable to the COMPAS dataset results in the main draft, showing that we can have more experts with smaller fairness constraint increments to produce a similar FEAMOE model.

\begin{table*}[h!]
\centering
\begin{tabular}{cc} 
\hline
Performance &   $k=40$ \\
\hline
Accuracy   &  63.77   \\
SPD      &  0.05  \\
AOD    & 0.13    \\
Burden    & 0.05  \\
\hline
\end{tabular}
  \caption{Results on the COMPAS dataset with $k=40$ and fairness increments of 0.002} \label{Table1}

\end{table*}

\begin{figure*}[!t]
     \centering
     \begin{subfigure}[b]{0.45\textwidth}
         \centering
         \includegraphics[width=\textwidth]{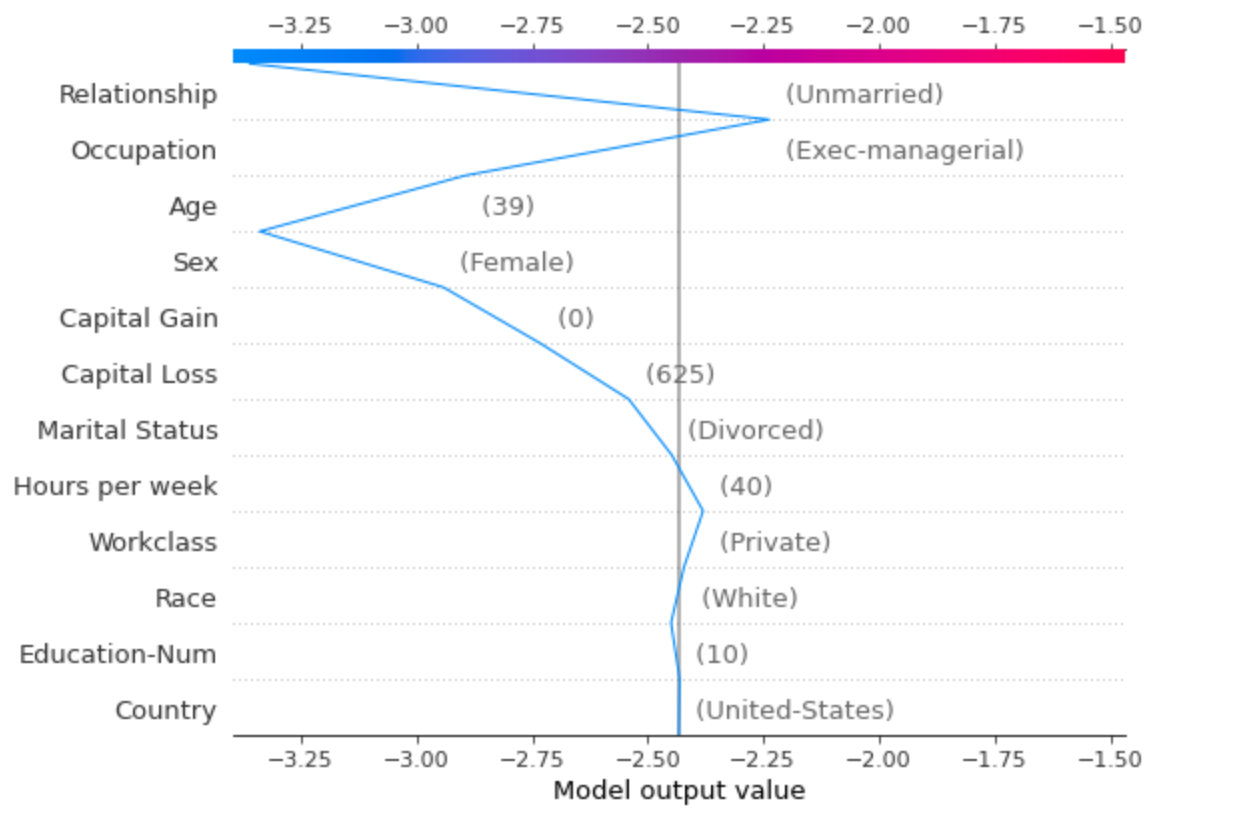}
         \caption{}
         \label{fig:a}
     \end{subfigure}
     \hfill
     \begin{subfigure}[b]{0.47\textwidth}
         \centering
         \includegraphics[width=\textwidth]{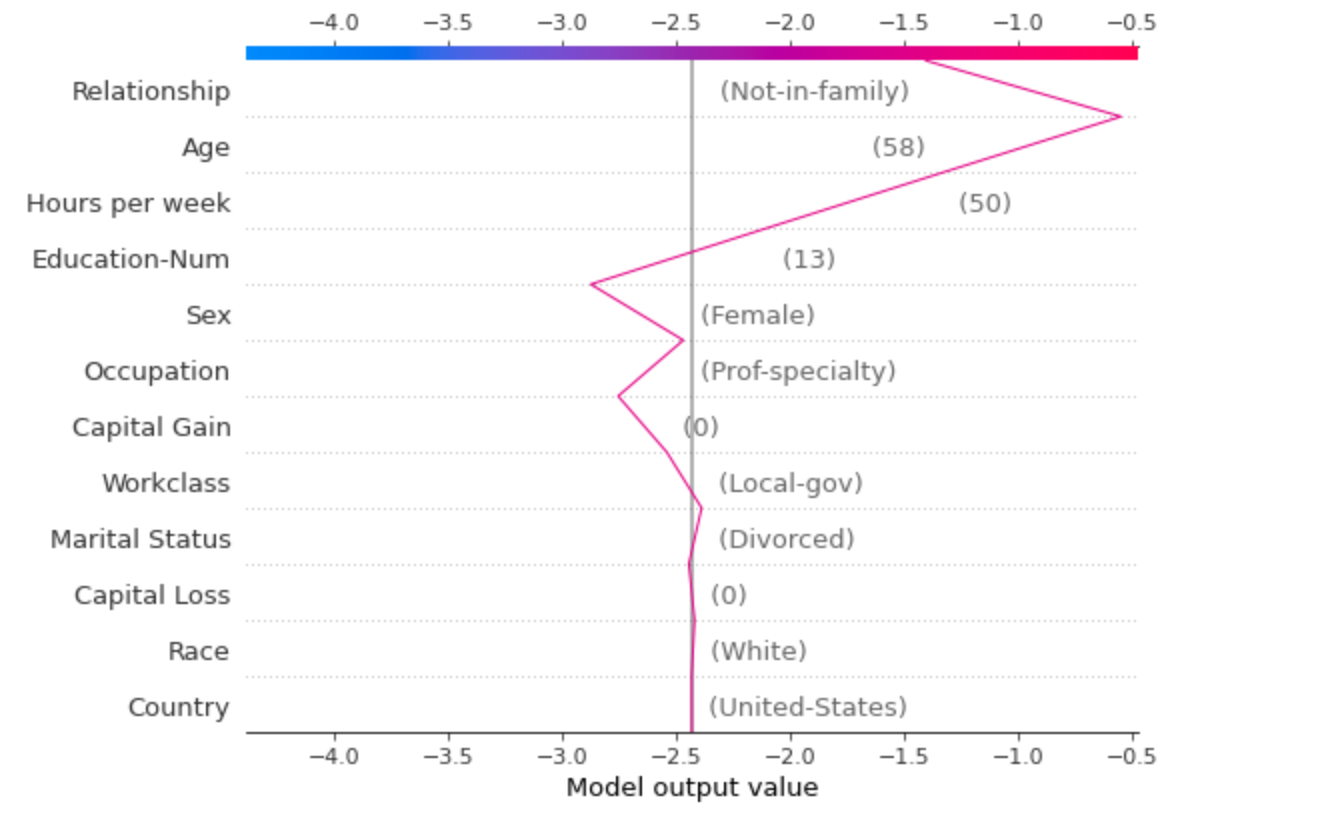}
         \caption{}
         \label{fig:b}
     \end{subfigure}
        \caption{Shapley value explanations for two different data points using the mixture of experts model by finding independent Shapley values and then using Theorem 1. KernelShap for a neural network model by using all feature combinations showed similar results, while taking much longer to compute.}
        \label{fig:Shapexp}
\end{figure*}

\subsection{Example Shapley value explanations}

Some example Shapley value explanations based decision plots for individual test points for the UCI Adult dataset are shown in Figure \ref{fig:Shapexp}. We find independent Shapley values for data points for every expert using the setup provided in https://slundberg.github.io/shap/notebooks/plots/decision\_plot.html and then use Theorem 1 to find the final Shapley values. KernelShap used on neural networks for the same data points had similar results, while taking significantly longer to compute


\end{document}